\documentclass[letterpaper, 10 pt, conference]{ieeeconf} 
\IEEEoverridecommandlockouts                  
\usepackage{graphicx}
\usepackage{mathtools}
\usepackage{amsmath}
\usepackage[cmintegrals]{newtxmath}
\usepackage{url}
\usepackage[usestackEOL]{stackengine}
\usepackage{subfig}
\usepackage{color}
\usepackage{algorithm}
\usepackage[noend]{algpseudocode}
\usepackage{balance}

\title{\LARGE \bf{Mobile Robot Localisation and Navigation Using\\ LEGO NXT and Ultrasonic Sensor}}
\author{Yanan Liu$^{1}$,~\IEEEmembership{Graduate Student Member,~IEEE,}~Rui Fan$^{2}$,~\IEEEmembership{Member,~IEEE,} Bin Yu$^{1}$, \\ M. Junaid Bocus$^{3}$,~\IEEEmembership{Graduate Student Member,~IEEE,} Ming Liu$^{2}$,~\IEEEmembership{Senior Member,~IEEE,} Hepeng Ni$^{4}$, \\Jiahe Fan$^{5}$,~\IEEEmembership{Student Member,~IEEE,} Shixin Mao$^{6}$
\thanks{*This work was supported by China Scholarship Council (CSC) (No. 201700260083), the Bristol Robotics Laboratory, and Visual Information Laboratory.}
\thanks{$^{1}$Bristol Robotics Laboratory, University of Bristol, Bristol, United Kingdom. {\tt\small yanan.liu@bristol.ac.uk}}
\thanks{$^{2}$Robotics and Multi-Perception Laboratory, Department of Electronic and Computer Engineering, The Hong Kong University of Science and Technology, Hong Kong.}%
\thanks{$^{3}$Department of Electrical and Electronic Engineering, University of Bristol, Bristol, BS8 1UB, United Kingdom.}%
\thanks{$^{4}$School of Mechanical Engineering, Shandong University, Jinan, 250061, China.}%
\thanks{$^{5}$School of Communication Engineering, Jilin University, Changchun, 130000, China.}%
\thanks{$^{6}$Shenzhen Sanhang Industrial Technology Research Institute, Shenzhen, 518000, China.}%
}

\begin{document}
\maketitle

\begin{abstract}
Mobile robots are becoming increasingly important both for individuals and industries. Mobile robotic technology is not only utilised by experts in this field but is also very popular among amateurs. However, implementing a mobile robot to perform tasks autonomously can be expensive because of the need for various types of sensors and the high price of robot platforms. Hence, in this paper we present a mobile robot localisation and navigation system which uses a LEGO ultrasonic sensor in an indoor map based on  the LEGO MINDSTORM NXT. This provides an affordable and ready-to-use option for most robot fans. In this paper, an effective method is proposed to extract useful information from the distorted readings collected by the ultrasonic sensor. Then, the particle filter is used to localise the robot. After robot's position is estimated, a sampling-based path planning method is proposed for the robot navigation. This method reduces the robot accumulative motion error by minimising robot turning times and covering distances. The robot localisation and navigation algorithms are implemented in MATLAB. Simulation results show an average accuracy between 1 and 3 cm for three different indoor map locations. Furthermore, experiments performed in a real setup show the effectiveness of the proposed methods.

\textit{Index terms}---mobile robot, LEGO NXT, localisation and navigation, particle filter, path plan.

\end{abstract}

\IEEEpeerreviewmaketitle

\section{Introduction}

Mobile robot localisation and navigation technologies play an important role in modern-day robots, which are being increasingly used in industry, driver-less cars, assisted living, logistics and domestic applications, especially in an ageing society \cite{Wang2017}. The state-of-the-art mobile robot technologies have been significantly enhanced with the furthered development of sensors and algorithms. Consequently, mobile robots are utilised in a wider range of applications than ever before. Companies such as JD.com are testing motorised robots for delivery purposes in universities \cite{huang2015robotics}. Moreover, Google and Baidu are developing a new generation of autonomous driver-less cars \cite{lipson2016driverless}. Automatic cleaning robots are being increasingly used in households nowadays \cite{gutmann2012social}. People are fascinated by the rapid development of mobile robots and hence a lot of research is being devoted to this field to make our lives better. Unfortunately, most of the advanced sensors are expensive for individuals. Hence, in this paper, we demonstrate an affordable and easy-to-use option for robot amateurs and allow them to develop a mobile robot system using the LEGO NXT and an ultrasonic sensor.

The LEGO NXT \cite{ranganathan2008use} is an off-the-shelf tool-kit to verify mobile robot control algorithms. It encompasses a new generation of programmable, educational \cite{Karp2010} and modularised robotics, enabling developers to easily understand, design and program robots \cite{Grega2008}.


A number of mobile robotic systems have been developed based on the LEGO NXT with various different types of sensors \cite{Pinto2012}. Lee and Buitrago \cite{Lee2015} proposed the map generation and robot localisation system which simply uses a PC camera and an ultrasonic sensor. However, the accuracy of robot positioning can be largely affected by a change in illumination in the environment. Consequently, it is unlikely to achieve robust operation. A Radio-Frequency IDentification (RFID)-based localisation framework was put forward by Chawla and Robins, which was realised using the LEGO NXT kits \cite{Chawla2011}. However, this solution requires putting tags in the environment for the readers to track. Therefore, the localisation accuracy depends on the position of the tags to some extent. Moreno et al. \cite{Moreno2002} and Li et al. \cite{Lee2010} also achieved robot localisation and navigation with several or several types of sensors. Thus, these solutions, where many sensors are used, would be more expensive.

In this paper we utilise a LEGO ultrasonic sensor as the only perception tool which performs both the localisation and navigation tasks. This method not only makes the robot system inexpensive but also simplifies the software architecture. However, the ultrasonic sensor data is often mixed with propagation phenomena such as noise, echoes, reflections, attenuation and other limiting factors \cite{Garc2004}. Consequently, to obtain an accurate estimation of the robot position, the sensor readings should be manipulated in such a way that useful information can be extracted from the noisy and distorted data. Hence, in this paper, a practical method is proposed to process the acoustic data for localisation purposes. The processed sensor data acts as an input for the particle filter method, which probabilistically estimates the robot position \cite{Gustafsson2010}. The particle filter approach is widely applied to solve uncertainty problems based on probability theory and the Bayesian theorem \cite{Nurminen2013}. Once the robot's position is probabilistically estimated, a path to the target position is planned with the sampling-based method \cite{Karaman2011}. Unlike the popular A* path planning method, which produces a gridded trajectory and usually causes localisation error in a gridded decomposed map \cite{Ganeshmurthy201502}, this paper uses sampling based path planning method to select optimal sample points and form a short, collision-free trajectory with minimal turning times. A re-localisation and re-navigation mechanism is designed in case the robot is positioned incorrectly.

\begin{figure}[t!]
\centering
\parbox{3.3in}{\includegraphics[width=3.3in]{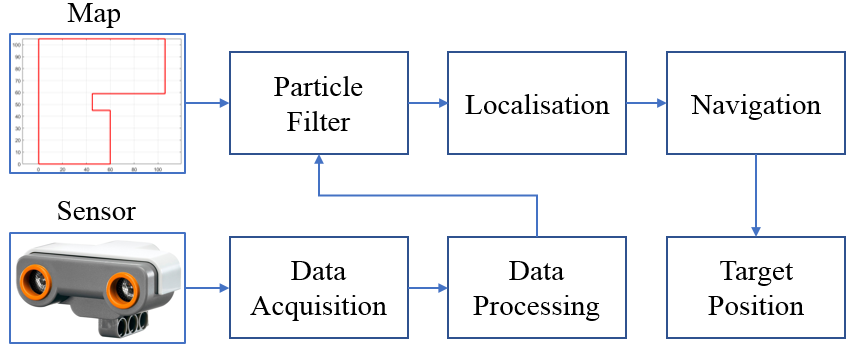}}
\caption{Overview of the proposed robot control process.}
\label{fig.overview}
\end{figure}

Finally, all the robot control algorithms are implemented using the RWTH Mindstorms NXT Toolbox for MATLAB and these are run on a computer that connects to the robot through a USB cable. An overview of the proposed robot control process is illustrated in Fig. \ref{fig.overview}.






The remaining sections of this paper are organised as follows: Section II describes the proposed robot control system including its mechanical structure, the robot's functionalities and the algorithm's description. In Section III, the simulation and experimental results are provided and the proposed system performance is evaluated. Finally, Section IV summarises the paper and provides recommendations for future work.


\section{System Description}
This section describes the robot's mechanical structure, software functionalities, localisation algorithms and path planning algorithms.

\subsection{Robot Design}

\begin{figure}[t!]
\centering
\parbox{3in}{\includegraphics[width=3in]{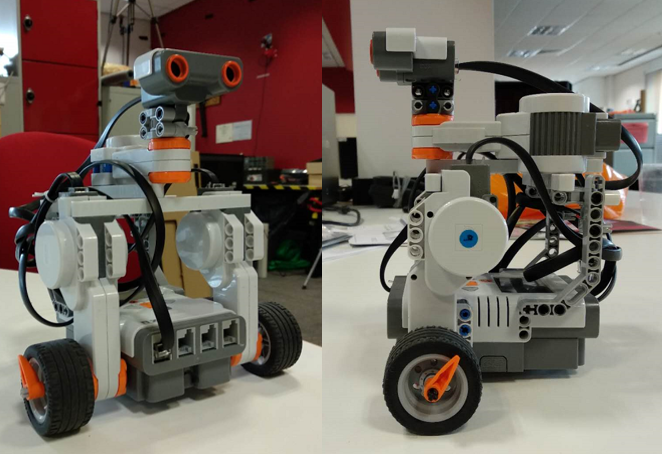}}
\caption{The mechanical structure of the mobile robot.}
\label{lego robot}
\end{figure}

In the initial phase, we aimed to design a robot which has a simple, light and easily controlled structure. The differential drive structure is appropriate for indoor robot navigation because it enables the robot to flexibly change direction in a confined place. Furthermore, only two motors are needed to drive the robot, allowing it to be light-weight. The third servo motor can rotate the ultrasonic sensor 360$^\circ$ around the motor axis, enabling the sensor to get adequate readings from the environment. The rotating sensor is installed in such a way that it is homocentric to the robot's centre of motion to simplify the geometrical complexity. Moreover, the position obtained from the particle filter can be directly used as the robot motion centre without any further coordinate transformations. Finally, the height of the sensor must not be greater than the height of the wall. The robot hardware architecture is shown in Fig. \ref{lego robot}.

\subsection{Some Basic Robot Functionalities}
Based on the existing NXT library, we developed a number of function modules that can be easily called in MATLAB for the robot's motion control and sensor operation. These basic functions are described as follows:

\subsubsection{Move}
This function has three main features: 

\begin{itemize}
\item Move the robot forward by a given distance in the same direction it is facing.
\item Force the robot to stop when it gets too close to the wall.
\item Output a boolean value indicating whether there is a potential collision with the wall.
\end{itemize}

\subsubsection{Turn}
This function allows the robot to rotate itself around its centre of motion which is the midpoint between its two wheels. 

\subsubsection{Turnback}
This function is always running when the robot is performing its tasks. If the distance between the wall and the robot is less than a given threshold, this function will force the robot to stop and then instruct it to move the robot backwards by a certain distance to maintain a safe distance from the wall.

\subsubsection{Ultrascan}
This function instructs the ultrasonic sensor to perform a quick 360$^\circ$  rotation around the servo axis to collect readings. During the rotation, the ultrasonic sensor collects a series of distances to the wall with respect to different angles. Then, the ultrasonic sensor rotates back to its original position and starts to detect potential objects in front of it.

\subsection{Algorithm Description}
In this section, we first analyse the characteristics of the ultrasonic sensor by comparing the real data and sensor readings. Then, a data processing method is proposed to handle the sensor data. Finally, we describe the particle filter localisation method and a sampling-based path planning method.

\subsubsection{Characteristics of the Ultrasonic Sensor}

\begin{figure}[t!]
\centering
\parbox{2.5in}{\includegraphics[width=2.5in]{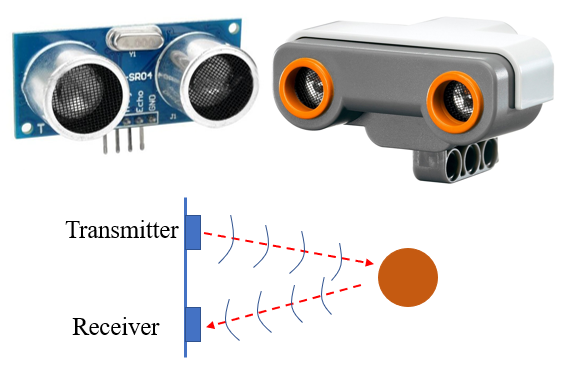}}
\caption{LEGO ultrasonic sensor.}
\label{ultrasonic sensor}
\end{figure}

Ultrasonic sensors are among the most popular sensors used for robot indoor localisation because they are cheap, lightweight, compact and have low energy consumption \cite{Zhao2012}. Consequently, this paper uses a LEGO ultrasonic sensor (Fig. \ref{ultrasonic sensor}) for the purpose of localisation and navigation tasks.

As an active sensor, the ultrasonic sensor emits acoustic waves that propagate through the environment and are then captured by the receiver. The distance travelled by the acoustic waves \textit{d}, can be calculated as follows:

\begin{equation} 
 d = c*t/2 
\end{equation}
where \textit{c} is the propagation velocity of the acoustic waves and \textit{t} is the time taken for the acoustic waves to reach the receiver.

However, the ultrasonic sensor is not accurate enough to measure the time of flight \textit{t} \cite{Garc2004}. Another major limitation of the ultrasonic sensor is that the ultrasonic waves form a cone with a certain opening angle when propagating through the environment \cite{siegwart2011introduction}. As a result, ultrasonic sensors cannot feedback accurate depth information for some orientations.

\begin{figure}[t!]
    \centering
    \subfloat[Robot position in the map, yellow points are cross points of scanning lights and wall. The red line is the boundary of the map.]{{\includegraphics[width=4.5cm]{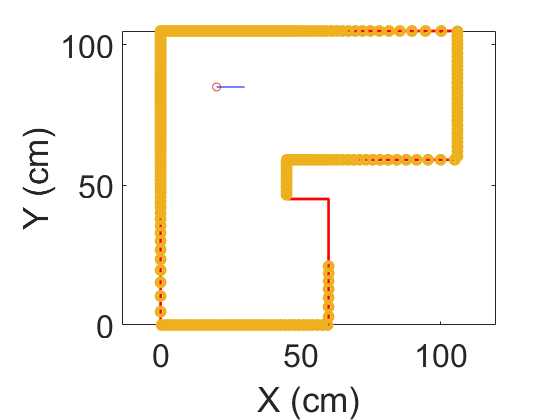} }}%
    \,
    \subfloat[The ultrasonic sensor scan, red lines are formed by simulated sensor readings, blue lines are formed by real sensor readings.]{{\includegraphics[width=3.7cm]{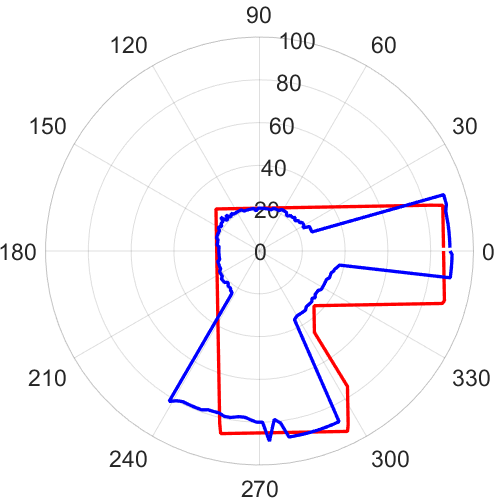} }}%
     \,
    \subfloat[Intersection points when the robot is at position (86, 85).]{{\includegraphics[width=4.5cm]{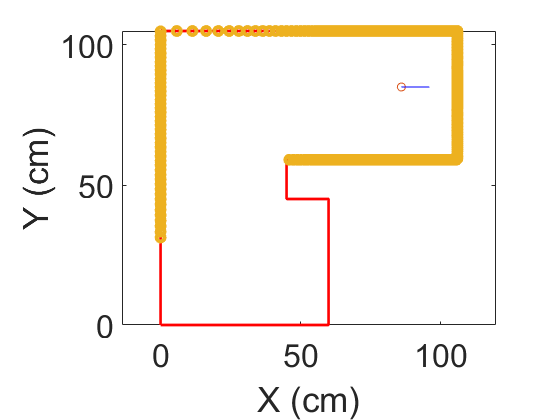} }}%
     \,
    \subfloat[Sensor readings when robot is at position (86, 85).]{{\includegraphics[width=3.7cm]{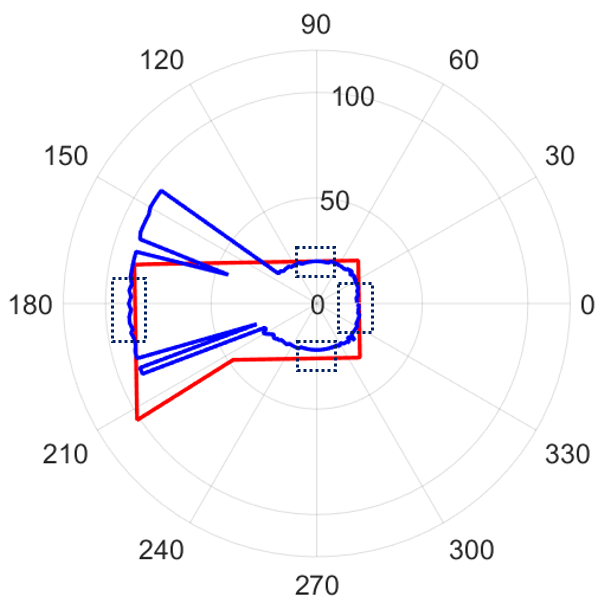} }}%
    \caption{Scannings of a typical ultrasonic sensor in the arena.}%
    \label{scannings}%
\end{figure}

Before applying the ultrasonic scan data of the robot to perform particle scoring, we plotted it on a radar chart to verify its accuracy. Fig. \ref{scannings} illustrates the inaccuracy of the readings when scanning a map with an ultrasonic sensor. As shown in Fig. \ref{scannings} (a) and (b), there is a notable inconsistency between the scanning data of a simulated particle and real robot. The robot starts scanning at a position (20, 85) with an orientation of 0$^\circ$ in the map. The real sensor readings are distorted because of wave reflections from the wall and the dispersal of the wave cone. Furthermore, sensor reading errors occur due to echoes, attenuation and other uncertain factors. Consequently, the raw scanning data is not suitable to localise the robot directly. 

\subsubsection{Ultrasonic Sensor Data Processing}
After further investigation, we found that the inaccuracy increased dramatically as the angle between the wave propagation and the wall increases (Fig. \ref{scannings} (b) and (d)). Therefore, we take the minimum of the scanning data and turn our robot to that direction so that ultrasonic sensor will be approximately perpendicular to the wall. Then, we keep the scanning data within the range of $\pm15$$^\circ$ at 0$^\circ$, 90$^\circ$, 180$^\circ$ and 270$^\circ$ positions so that we can ensure the accuracy while keeping enough data for particle filter. As shown in Fig. \ref{scannings} (d), the scanning data in the black dotted box can be used for localisation.

\subsubsection{Particle Filter Localisation}

For the localisation part, we utilise particle filter to probabilistically estimate the location of the robot in the map. The particle filter approach is widely applied to solve uncertainty problems based on probability theory and Bayesian theorem. It has a number of successful applications filed in robotics  \cite{fan2018real, Fan2018, Fan2016, Ozgunalp2017, Fan2018a}. In this scenario, the uncertainty of the robot location and orientation (\textit{$x, y, \theta$}) is represented by particle distributions. 

To localise a robot, we first fill the given map with randomly distributed particles where each particle potentially represents the real robot position and orientation. Then, the robot sensor readings and simulated particle reading are compared to calculate each particle's probability to be the true position. After ranking the particles according to their probability, we reallocate the particles around previously generated particles with higher weights. By doing this iteratively, the scope of the robot position can be gradually narrowed down.

Particle filter localisation is developed based on Bayesian theorem (Eq. \ref{eq:2}) as follows:
\begin{equation} 
P(A|B) = \frac{P(B|A) \, P(A)}{P(B)} \label{eq:2}
\end{equation}
where, $P(A | B)$ represents the probability that the particle position is the robot position, given a particle reading \textit{B}. \textit{P(A)} is the prior knowledge which equals to $P(A | B)$ in the previous step. Furthermore, \textit{P(B)} is an independent parameter which can be obtained from the particle scanning. Finally, $P(B | A)$ represents the likelihood of the particle reading \textit{B}, given a robot position \textit{A}. Apparently, $P(B | A)$ can not be obtained directly. However, it can be known by taking advantage of the actual sensor readings \textit{R} instead of the unknown robot position \textit{A} \cite{thrun2005probabilistic},

\begin{equation} 
P(B|A) \propto P(B|R) \label{eq:3}
\end{equation}

Moreover, the ultrasonic sensor model is based on a normal distribution given by Gaussian function \cite{thrun2005probabilistic}. Applying Eq. \ref{eq:3}, $P(B | A)$ can be obtained as follows:

\begin{equation} 
P(B|A) \propto P(B|R) \propto \frac{1}{{\sigma \sqrt {2\pi } }}e^{{{ - \left( {b - r } \right)^2 } \mathord{\left/ {\vphantom {{ - \left( {b - r } \right)^2 } {2\sigma ^2 }}} \right. \kern-\nulldelimiterspace} {2\sigma ^2 }}}  \label{eq:4}
\end{equation}






\subsubsection{Sampling-Based Path Planning}

Once the robot position is probabilistically estimated, a collision free path needs to be planned. For this scenario, the turning error of the robot impacts on the final position error. Moreover, low energy consumption is favoured for mobile robot. Consequently, we utilise the sampling-based method to plan a path with minimum turning times and short distance. The proposed method is described in Algorithm \ref{alg:pathplan}.

\begin{algorithm}
\caption{Sampling based path planning method}\label{alg:pathplan}
\begin{algorithmic}[1]
\State calculate the configuration space: \textit{C}
\State distribute \textit{N} particles in \textit{C} randomly
\State current path distance \textit{l} = $\infty$
\While{$i\not=\textit{N}$}
\State $i++$
\State select \textit{i} particles
\State link start point, \textit{i} particles and end point with straight lines
\If{ no collisions detected $and$ path length $Pi \textless l$}
  \State $ l = Pi$
\EndIf
\If{ a path found with $i$ particles}
   \State break
\EndIf
\EndWhile
\State \textbf{end}
\end{algorithmic}
\end{algorithm}

At first, a number of particles are distributed in the configuration space randomly. With this method, a collision free path can be guaranteed. Particles are selected according to the particle numbers and path length. The lesser the number of particles, the fewer the number of turning times for the robot. 

\begin{figure}[t!]
\centering
\parbox{2.8in}{\includegraphics[width=2.8in]{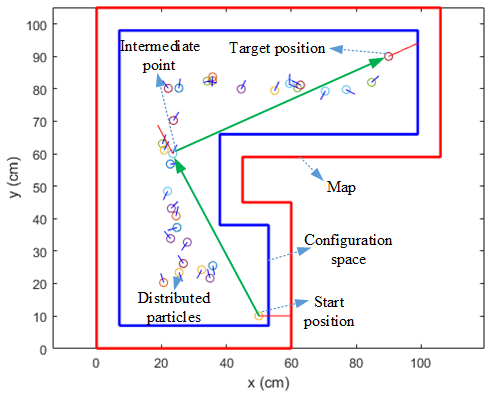}}
\caption{Sampling based path planning.}
\label{fig.path planning}
\end{figure}

As shown in Fig. \ref{fig.path planning}, the robot starts at (50, 10) and the target position is (90, 90). To prevent collision, we included an offset which is half the mobile length inside the original map to produce a configuration space. Then, 30 intermediate waypoints are distributed in the configuration space randomly. A safe path consisting of several intermediate waypoints(usually 0 - 2 for this type of map) can be found. Finally, in this case, one intermediate point is selected out of these 30 particles according to the length of the whole path.

A* path planning is a popular robot path planning method. However, it causes localisation error in a gridded decomposed map. Fig. \ref{fig.path planning} illustrates the effectiveness of the sampling based path planning method . The sampling based method allows for a shorter path length and minimum turning times, enabling the mobile robot to move towards the desired position with minimal turning errors and running time.

\begin{figure}[t!]
\centering
\parbox{2.3in}{\includegraphics[width=2.3in]{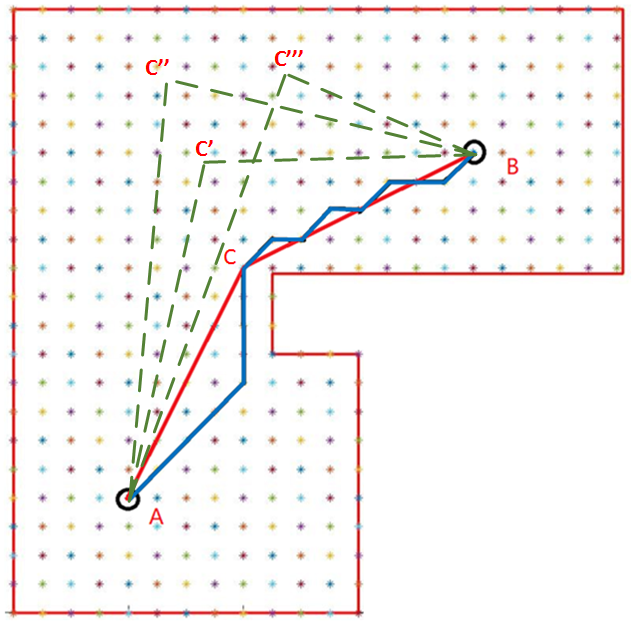}}
\caption{ A comparison between A* path planning (blue) and sampling based path planning (red). Start position A (20, 20), target position B (80, 80). Stars in map represent the grids of A* path planning. We assume there is a particle C distributed on a grid. According to the particle distribution situation, other possible sampling based paths are shown in dotted green lines.}
\label{path plan}
\end{figure}

However, when the robot is moving along the planned path, inaccurate robot position estimation can possibly cause robot to get very close to the arena wall, which may lead to a collision. To detect potential collisions, the ultrasonic sensor scanning is kept running at all times during the robot navigation process. If the distance between the robot and the wall is smaller than a given threshold, the robot will be instructed to move backwards and perform re-localisation and re-navigation. 


\section{Experimental Results}

\begin{table*}[h]
\caption{Localisation and navigation simulation results.}
\label{table.simulation result}
\begin{center}
\begin{tabular}{|c|c|c|c|c|}
\hline
 Map &Completion time (s)& Distance from target (cm) &  Path length (cm)& Detected collision\\
\hline
Map1 & 15.22 & 1.79 & 116.9 & 0\\
Map2 & 16.69 & 1.83 & 120.1 & 0\\
Map3 & 23.48 & 2.11 & 276.2 & 0\\
\hline
\end{tabular}
\end{center}
\end{table*}

In this section, we evaluate the performance of the proposed robot localisation and path planning algorithms both in terms of simulations and real arena setup. Experiments are carried out on a desktop computer having a Core i7 6700 CPU. The desktop PC and the robot controller NXT communicate with each other through a USB cable. The robot control algorithms run on the BotSim Robot simulator which was developed by the University of Bristol.

\subsection{Simulation Results}

\begin{figure}[t!]%
    \centering
    \subfloat[Map 1.]{{\includegraphics[width=2.5cm]{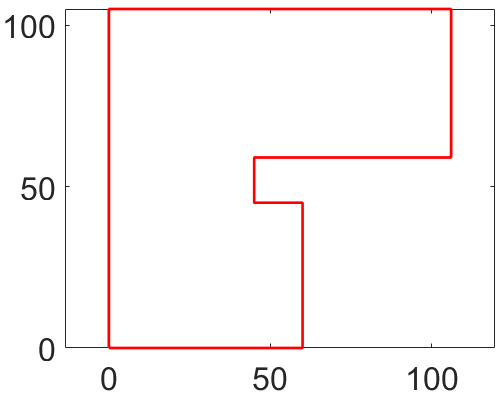} }}%
    \,
    \subfloat[Map 2.]{{\includegraphics[width=2.5cm]{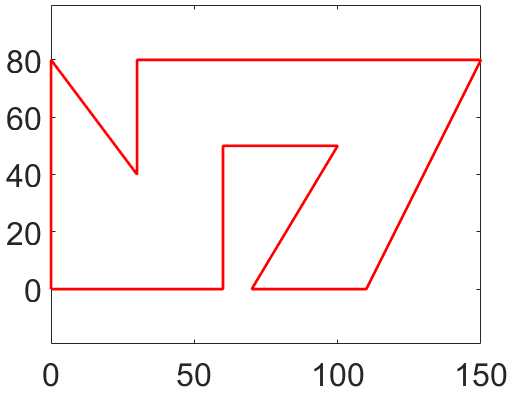} }}%
    \,
    \subfloat[Map 3.]{{\includegraphics[width=2.5cm]{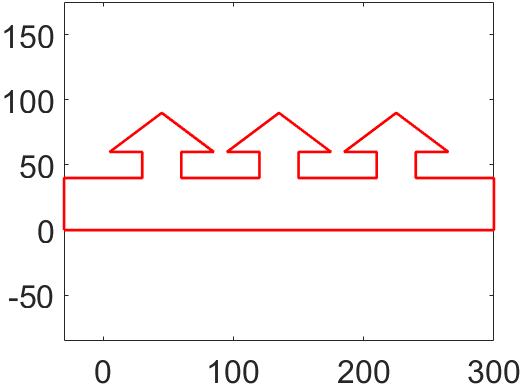} }}%
    \caption{Three different maps for simulation.}%
    \label{fig.three maps}%
\end{figure}

To evaluate the effectiveness of the proposed algorithms, we simulate the robot localisation and navigation algorithms in three different maps as shown in Fig. \ref{fig.three maps}. For this simulation, sensor scanning and robot motion noises are taken into consideration. The ultrasonic scanning error is set to 1 cm, while the robot motion error is 0.1 cm and the robot turning error is 0.005 radian. These parameters are set according to the actual performance of the sensor and robot in the environment. Tab. \ref{table.simulation result} shows the average test results for each map. Both the robot running time and target accuracy are recorded by the robot simulator. The final target accuracy is between 1 cm and 3 cm for a completion time between 15 and 24 seconds. Furthermore, no collision occurred, indicating  the robustness and effectiveness of the proposed sampling based path planning method.

\subsection{Experimental Results in the Arena}

\begin{figure}[t!]
\centering
\parbox{3.0in}{\includegraphics[width=3.0in]{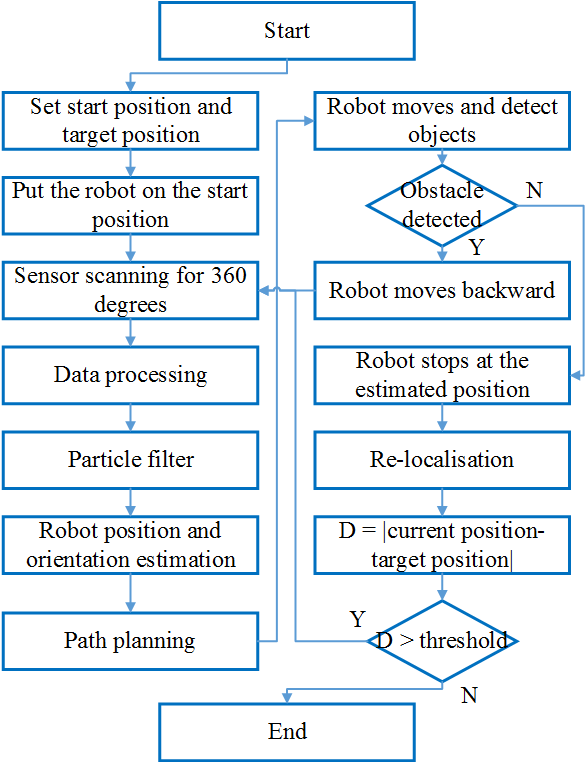}}
\caption{Robot operation process.}
\label{fig.robot operation process}
\end{figure}

\begin{figure}[t!]
\centering
\parbox{3.2in}{\includegraphics[width=3.2in]{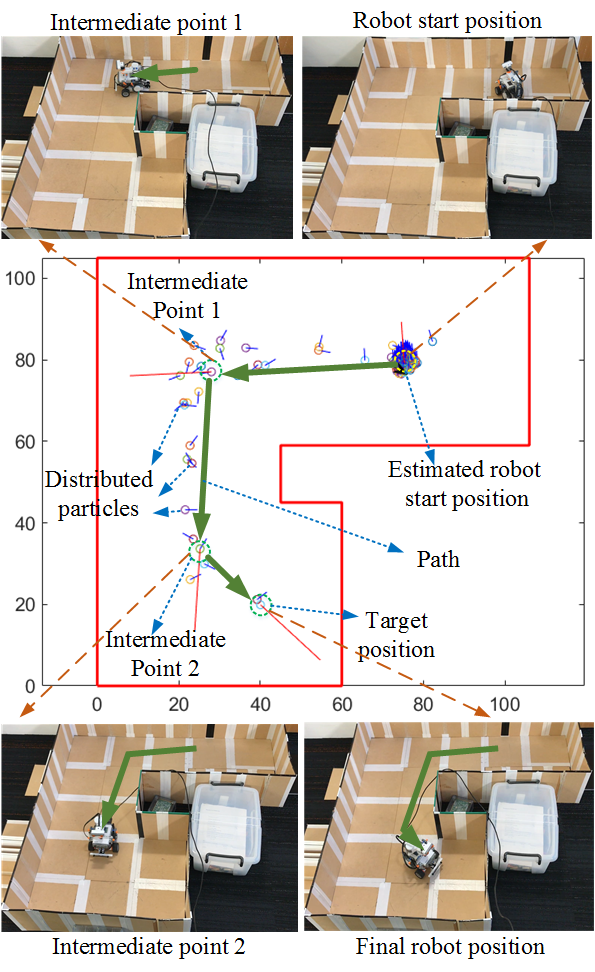}}
\caption{Robot localisation and navigation in the arena and the simulator.}
\label{fig.robot experiment v1}
\end{figure}

We conduct further real experiments in the arena (Map 1) to demonstrate the feasibility of our approaches. A detailed flowchart of the proposed robot control system is shown in Fig. \ref{fig.robot operation process}. A desktop PC running the MATLAB robot simulator is utilised to perform calculation and send instructions to the robot. At first, the robot starting position is randomly set. Then, the ultrasonic sensor is driven to scan the inside of the arena for 360$^{\circ}$. An ultrasonic reading is recorded for each unit angle. The estimation of robot position and orientation is obtained using the processed sensor data, which usually takes several seconds. Based on the estimated robot position and orientation, a collision free path is planned by linking several waypoints which are selected from the randomly-distributed intermediate particles. The mobile robot moves to each intermediate point in sequence until it reaches its final destination. A wrong position estimation would most possibly cause collisions. In this case, localisation and path planning need be performed again. Finally, if the robot reaches its destination, it would carry out re-localisation to ensure it is as close as possible to the correct target position.

The arena is made up of cardboard and it has the same dimensions as Map 1 in the simulator. As observed in Fig. \ref{fig.robot experiment v1}, the simulator calculates the start position based on the sensor readings and then selects the best intermediate points for the mobile robot. The vehicle successfully navigates through the map without collision and reaches the final target according to the simulator instructions. Fig. \ref{fig.robot experiment v1} shows the simulator process and the robot navigation process. The robot motion process follows the simulator's instructions precisely. Therefore, the effectiveness and the performance of the robot localisation and path planning system is successfully demonstrated by the experimental results.

\section{Conclusion and Future Work}

In this paper, we presented a mobile robot localisation and navigation system using the LEGO NXT in a map. The robot navigation task, which is carried out with only one ultrasonic sensor, has been proved to be feasible. Furthermore, an effective method has been proposed to deal with noisy data obtained from the ultrasonic sensor. The robot's position and orientation have been estimated using the particle filter algorithm. In addition, a sampling based robot path was planned with minimum turning times and short path distance. Simulation results show that for the simulation in Map 1, a final target accuracy of 1.79 cm was obtained for a completion time of 15.22 seconds. Finally, the accuracy of our algorithms was backed by tests in a real arena.

One of the downsides of our work is that we could not effectively guarantee the robustness of the system in every real map. In future contributions, we aim to improve the system adaptability for different types of maps.


\end{document}